\documentclass[times,12pt]{article}

\usepackage{times} 
\usepackage{epsfig} 
\usepackage{amssymb}  
\usepackage{latexsym}   
\usepackage{epsfig}   
\usepackage{color}
\usepackage{verbatim}

\usepackage{rotating}
\def\rotleq{\,$\raisebox{-0.3mm}{\begin{turn}{90}$\leq$\end{turn}}$\,}

\def\bR{\begin{color}{red}} 
\def\bB{\begin{color}{blue}}
\def\bM{\begin{color}{magenta}}
\def\bC{\begin{color}{cyan}}
\def\bW{\begin{color}{white}}
\def\bBl{\begin{color}{black}} 
\def\bG{\begin{color}{green}}
\def\bY{\begin{color}{yellow}}
\def\e{\end{color}}

\newcommand{\bit}{\begin{itemize}}
\newcommand{\eit}{\end{itemize}\par\noindent}
\newcommand{\ben}{\begin{enumerate}}
\newcommand{\een}{\end{enumerate}\par\noindent}
\newcommand{\beq}{\begin{equation}}
\newcommand{\eeq}{\end{equation}\par\noindent}
\newcommand{\beqa}{\begin{eqnarray*}}
\newcommand{\eeqa}{\end{eqnarray*}\par\noindent} 
\newcommand{\beqn}{\begin{eqnarray}}  
\newcommand{\eeqn}{\end{eqnarray}\par\noindent}

\title{A quantum teleportation inspired algorithm\\  produces sentence meaning from word meaning and  grammatical structure} 

\author{
Stephen Clark\\ 
{\em University of Cambridge}\\
{\em  Computer Laboratory}\\ \ \\
Bob Coecke, Edward Grefenstette, Stephen Pulman\\
{\em University of Oxford}\\ 
{\em Department of Computer Science}\\ \ \\
Mehrnoosh Sadrzadeh\\
{\em Queen Mary, London}\\ 
{\em School of Electronic Engineering and Computer Science}}   
\date{}

\begin{document}  

 \maketitle  

\begin{abstract}
We discuss an algorithm which produces the meaning of a sentence given meanings of its words, and its resemblance to quantum teleportation. In fact, this protocol was the main source of inspiration for this algorithm which has many applications in the area of Natural Language Processing. \\
\end{abstract}

Quantum teleportation \cite{BBC} is  one of the most conceptually challenging and practically useful concepts that has emerged from the quantum information revolution.  For example, 
via logic-gate teleportation \cite{Gottesman} it gave rise to  the measurement-based computational model, it also plays a key role in current investigations into the nature of quantum correlations, e.g.~\cite{Popescu}, and it even has been proposed as a model for time travel \cite{BennettSchumacher}. It also formed the cornerstone for a new  axiomatic approach and diagrammatic calculus for quantum theory \cite{AC, ContPhys, NJP}.

Arguably, when such a radically new concept emerges in a novel foundational area of scientific investigation, one may expect that the resulting  conceptual and structural insights could also lead to progress in other areas, something which has happened on many occasions in the history of physics. In the context of quantum information, for example, it is well-known that quantum complexity theory has helped to solve many problems in classical complexity theory. 


Here we explain how a high-level description of quantum teleportation
with emphasis on information flows has successfully helped to solve a
longstanding open problem in the area of Natural Language Processing
(NLP), and the problem of modeling meaning for natural language more
generally \cite{CCS, CSC}.  This work featured as a cover
heading in the New Scientist (11 Dec.~2011) \cite{NewScientist},
and has been experimentally tested for its capability to perform key
NLP tasks such as word sense disambiguation in context
\cite{GS}.\footnote{EMNLP is the leading conference on corpus-based
  experimental NLP.}



\paragraph{The NLP problem.}  Dictionaries
explain the meanings of words; however, in natural language words are
organized as sentences, but we don't have dictionaries that explain
the meanings of sentences.  Still, a sentence carries more information
than the words it is made up from; e.g.~{\sf meaning(Alice sends a
  message to Bob) $\not=$ meaning(Bob sends a message to
  Alice)}. Evidently, this is where grammatical structure comes into
play.  Consequently, we as humans must use some algorithm that
converts the meanings of words, via the grammatical structure, into
the meaning of a sentence. All of this may seem to be only of academic
interest; however, search engines such as Google face exactly the same
challenge.  They typically read a string of words as a `bag of words',
ignoring the grammatical structure.  This is simply because (until
recently) there was no mathematical model for assigning meanings to
sentences.\footnote{More precisely, there was no mathematical model
  for assigning meanings to sentences that went beyond
  truthfulness. Montague semantics \cite{Montague} is a compositional
  model of meaning, but at most assigns truth values to sentences, and
  evidently there is more to sentence meaning than the mere binary
  assignment of either true or false.}  On the other hand, there is a
widely used model for word meaning, the \em vector space model \em
\cite{VectModel}.

 This vector space model of word meaning works as follows.  One
 chooses a set of context words which will form the basis vectors of a
 vector space.\footnote{These context words may include nouns, verbs
   etc.; the vector space model built from the British National Corpus
   typically contains 10s of thousands of these words as basis
   vectors.}  Given a word to which one wishes to assign meaning,
 e.g.~`Alice', one relies on a large corpus, e.g.~(part of) the web,
 to establish the relative frequency that `Alice' occurs `close' to
 each of these basis words.  The list of all these relative
 frequencies yields a vector that represents this word, its \em
 meaning vector\em.  Now, if one wants to verify synonymy of two
 words, it suffices to compute the inner-product of the meaning
 vectors of these words, and verify how close it is to 1. Indeed,
 since synonyms are interchangeable, one would expect them to
 typically occur in the context of the same words, and hence their
 meaning vectors should be the same in the statistical limit.  For
 example, in a corpus mainly consisting of computer science literature,
 one would expect Alice and Bob to always occur in the same context
 and hence their meaning vectors would almost be the same.  Of course,
 if the corpus were English literature (cf.~\cite{Alice}), then
 this similarity would break down.

Until recently, the state of affairs in computational linguistics was
one of two separate communities \cite{Gazdar}. One community focused on
\em non-compositional \em purely \em distributional \em methods such
as the vector space model described above. The other community studied
the \em compositional \em mathematical structure of sentences,
building on work by Chomsky \cite{Chomsky}, Lambek
\cite{Lambek1} and Montague \cite{Montague}. This work is mainly about
the grammatical structure of sentences; \em grammatical type calculi
\em are algebraic gadgets that allow one to verify whether a sentence
has a correct grammatical structure.


\paragraph{Caps, cups, and teleportation.} In \cite{AC},  a novel axiomatic framework was proposed to reason about quantum informatic processes, which admits a sound and faithful purely diagrammatic calculus \cite{ContPhys}; for some more recent developments we refer to \cite{NJP}. Ideal post-selected teleportation provides the cornerstone for the diagrammatic reasoning techniques, e.g.~here is the derivation of the general teleportation protocol where the $f$-label represents both the measurement outcome and  the corresponding correction performed by Bob \cite{ContPhys}:  
\[
\epsfig{figure=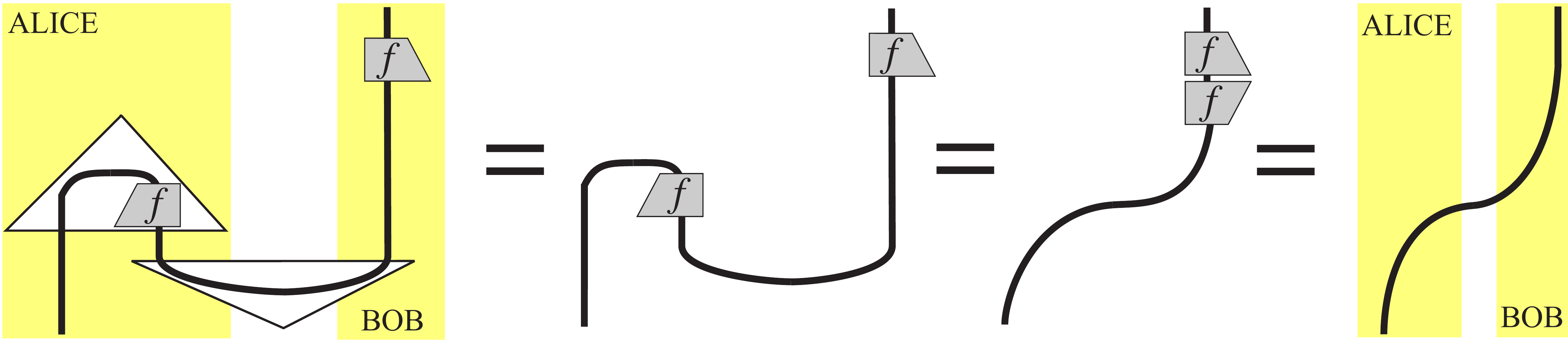,width=380pt}
\]
The main conceptual idea behind these diagrams is that, besides their operational physical meaning, they also admit  a `logical reading' in terms of \em information flow\em: 
\[
\epsfig{figure=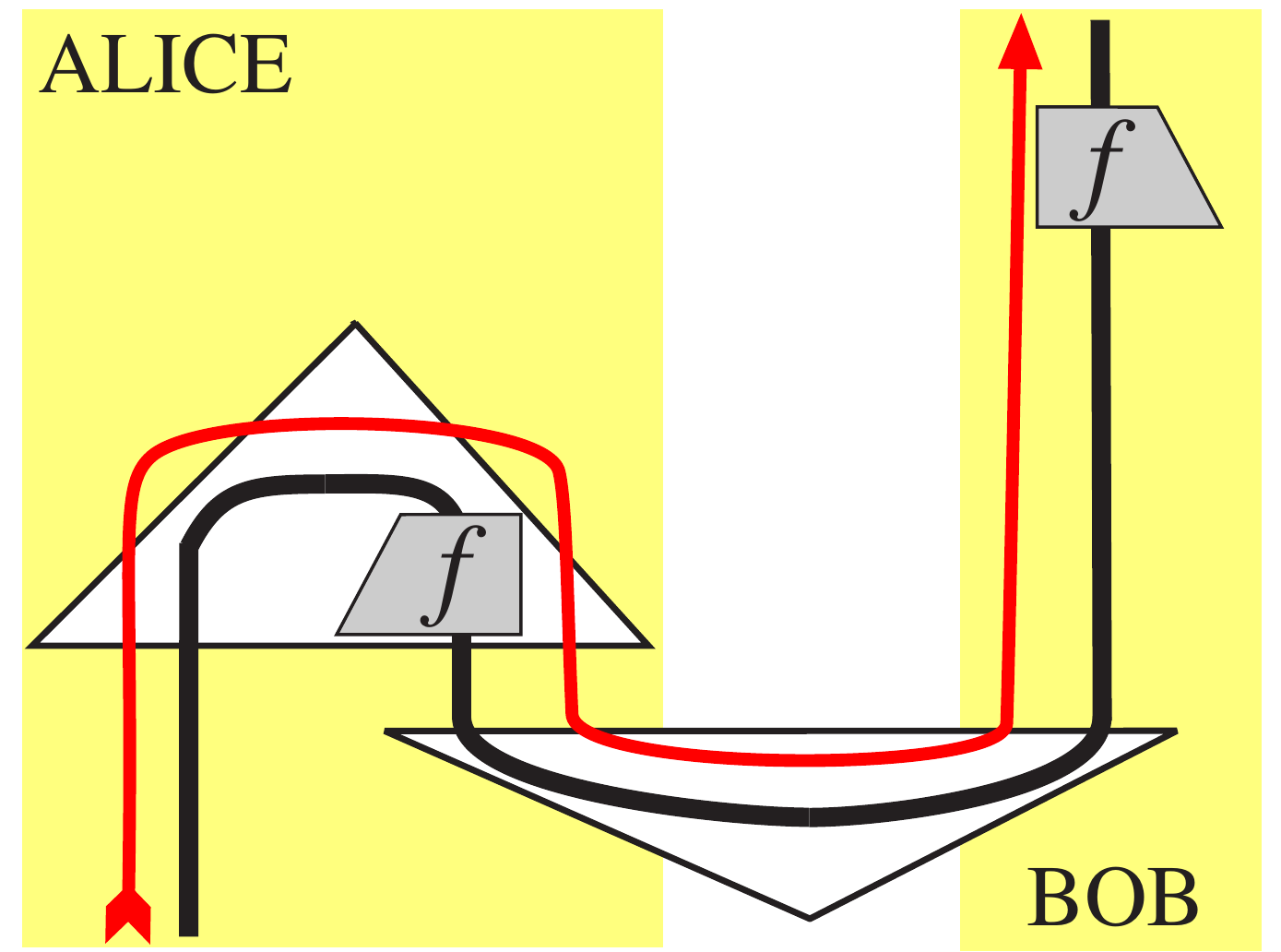,width=240pt}
\]
Here, the red line represents the logical flow which indicates that the state incoming at Alice's side first gets acted upon by an operation $f$, and then by its adjoint $f^\dagger$,  which in the case that $f$ is unitary results in the outgoing state at Bob's side being identical to the incoming one at Alice's side.\footnote{This `logical reading' of projectors on entangled states in terms of information flow was first proposed by one of the authors in \cite{LE}.} 

When interpreted in Hilbert space, the key ingredients of this formalism are `cups' and `caps':
\[
\raisebox{-4.8mm}{\epsfig{figure=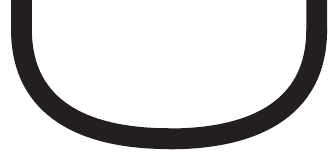,width=60pt}}:= |00\rangle +|11\rangle\qquad
\raisebox{-4mm}{\epsfig{figure=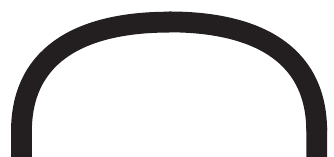,width=60pt}}:= \langle00|+\langle11|
\]
and the equation that governs them is:
\[
\bigl((\langle00|+\langle11|)\otimes Id\bigr) 
\bigl(Id\otimes(|00\rangle +|11\rangle)\bigr)=Id
\]
which diagrammatically depicts as:
\[
\epsfig{figure=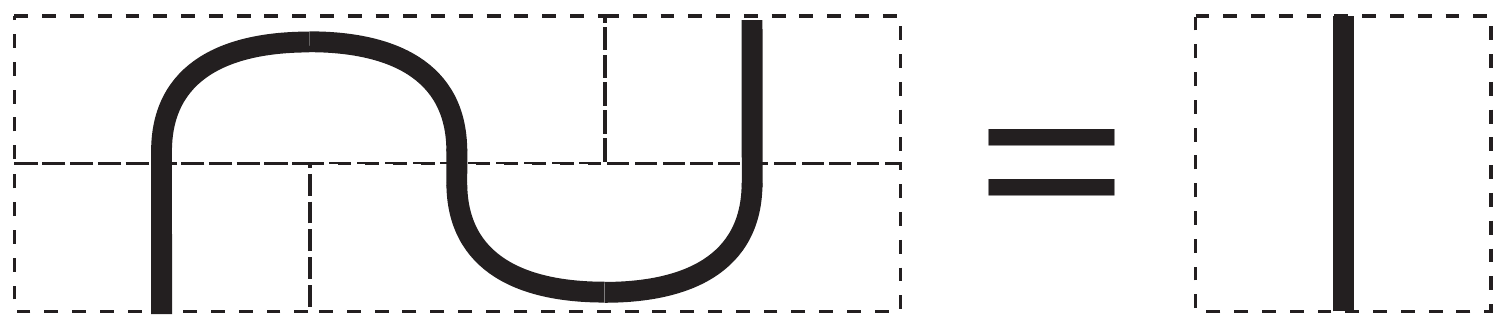,width=280pt}
\]
In this language the  Choi-Jamiolkowski isomorphism:
\[
|\Psi\rangle=\raisebox{-6mm}{\epsfig{figure=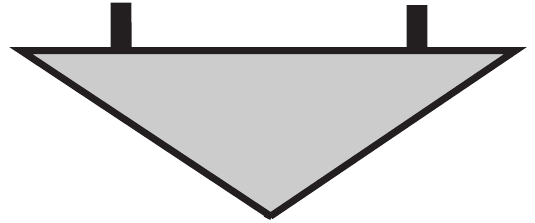,width=100pt}}\ =\raisebox{-6mm}{\epsfig{figure=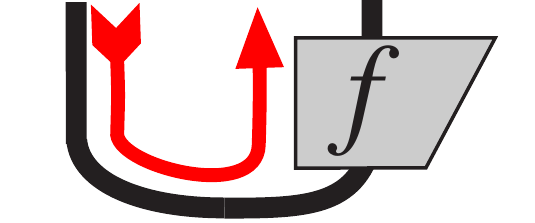,width=100pt}}\!\!\!=(Id\otimes f)(|00\rangle +|11\rangle)
\]
interprets a bipartite state (the grey triangle) as a `cup' which changes the direction of the information flow together with an operation $f$ that alters the information.  

Non-separatedness means  topological connectedness:
\[
|\Psi\rangle=\raisebox{-6mm}{\epsfig{figure=CJ1.pdf,width=100pt}}\ \not=\raisebox{-6mm}{\epsfig{figure=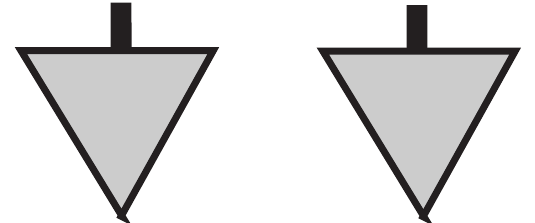,width=100pt}}=|\psi\rangle\otimes |\phi\rangle
\]
which is interpreted as the fact that information can flow between the two systems involved. Hence, when focussing on pure states, the cups effectively witness entanglement in terms of the information flows that it enables.  

It is exactly this interpretation of the vectors representing the states of compound quantum systems in terms of enabling information flows that will provide the cornerstone for our compositional and distributional model of meaning.

\paragraph{Solution to the NLP problem: the intuition.} Before we explain the precise algorithm that produces sentence meaning from word meaning, we provide the analogy with the above. 

A transitive verb requires both an object and a subject to yield a grammatically correct sentence. Consider the sentence ``Alice hates Bob''.  Assume that the words in it are represented by vectors, which as above we denote by triangles:
\[
\overrightarrow{Alice}\otimes \overrightarrow{hates}\otimes\overrightarrow{Bob}
=\raisebox{-6mm}{\epsfig{figure=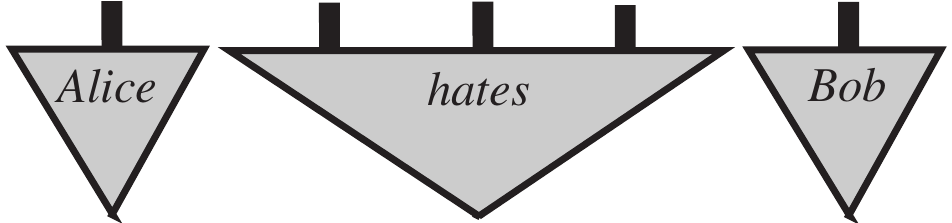,width=180pt}}
\]
Note here that treating verbs as `compound' was already the case in grammatical type calculi, as we discuss below.  So how do these words interact to produce the meaning of a sentence?  For the verb to produce the meaning of the sentence, that is, the statement of the fact that Alice hates Bob, it of course needs to know what its subject and object are, that is, it requires knowing their meanings.  Therefore, inspired by the above discussion on teleportation, we `feed' the meaning vectors $\overrightarrow{Alice}$ and $\overrightarrow{Bob}$ into the verb $\overrightarrow{hates}$ which then `spits out' the meaning of the sentence:
\[
\raisebox{-3mm}{\epsfig{figure=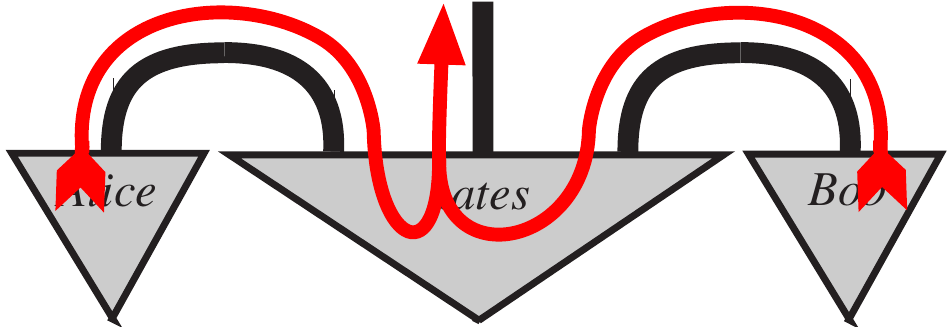,width=180pt}}
\]
Again, that the meaning of the sentence is produced by the transitive verb after interacting with its nouns is also something that was the case in grammatical type calculi.  

In the same vein, for an intransitive verb, we obtain an even more direct analogue to quantum teleportation:
\[
\left(\sum_i\langle ii|\otimes Id\right)(\overrightarrow{Alice}\otimes\overrightarrow{dreams})=\raisebox{-8mm}{\epsfig{figure=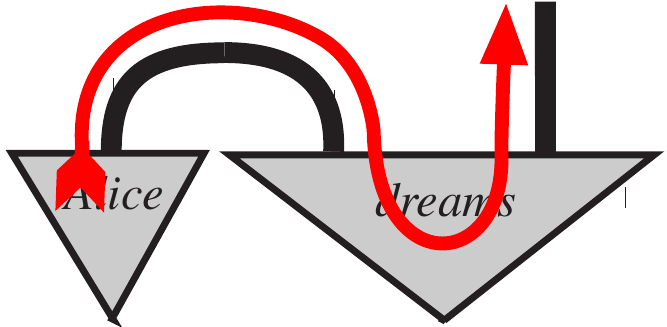,width=120pt}}
\]

Note here that non-separatedness of verbs is obvious: if in the sentence ``Alice hates Bob'' hates would be disconnected, then the meaning of the sentence would not depend on the  meanings of ``Alice'' and ``Bob'', so,  ``Anyone hates everyone''!

\paragraph{A grammatical type calculus: Lambek's pregroups.}
 In order to give a precise description of our algorithm we now give a brief account of Lambek's pregroup grammar \cite{Lambek2, Lambek3}.\footnote{An interesting aside: Lambek published his paper on the widely used Lambek grammars in 1958 \cite{Lambek1}. His recent book on pregroups appeared 50 years later \cite{Lambek3}!  All this time Lambek worked in Montreal, the  location of the upcoming QIP, and it was also in Montreal (in 2004) that during a talk by one of us he first pointed to  the structural coincidence between pregroup grammars and quantum axiomatics in terms of cups and caps.}

Pregroups capture structural similarities across a wide range of language families \cite{Lambek3}. 
They combine a remnant of group structure with partial ordering; the usual (left and right) group laws for the inverse are replaced by four inequalities involving distinct left  and right pseudo-inverses $x^{-1}$ and ${}^{-1}x$:
\[
x^{-1} \cdot x\leq 1\leq x \cdot x^{-1} \qquad x \cdot {}^{-1}x\leq 1\leq {}^{-1}x \cdot x\,.
\]
As a grammatical type calculus, its elements are basic grammatical types, e.g.~the noun type $n$ and sentence type $s$.  Other types arise from the pseudo-inverses and the multiplication, e.g.~the transitive verb type $tv:= {}^{-1} n\cdot s\cdot n{}^{-1}$. Then:
\beqa
\ \ \ \  \overbrace{n}^{\!\!\!\!\!\!\! e.g.~Alice\!\!\!}\cdot\overbrace{{}^{-1}n\cdot s\cdot n{}^{-1}}^{e.g.~hates}\cdot\overbrace{n}^{\!\!\!e.g.~Bob\!\!\!\!\!\!}
&=&(n\cdot {}^{-1}n)\cdot s\cdot (n{}^{-1}\cdot n)\\
&\leq& 1\cdot s\cdot 1\\
&=& s
\eeqa
Such an inequality $ n\cdot tv\cdot n\leq s$ then stands for the fact that ``noun -- transitive verb -- noun'' is a valid grammatical structure for a sentence.  Note here the correspondence with our interpretation of such a sentence in terms of information flow: the verb requires two nouns to be `fed  into it' to yield a sentence type; $n^{-1}$ and ${}^{-1}n$ capture `the verb requests type $n$ on the left/right'. In fact, 
the  inequalities using $n\cdot {}^{-1}n\leq 1$ and $n^{-1} \cdot n\leq 1$ can also be represented with `directed' caps:  \beq\label{diagtypered}
\begin{array}{c}
s\\
\raisebox{-1.5mm}{\epsfig{figure=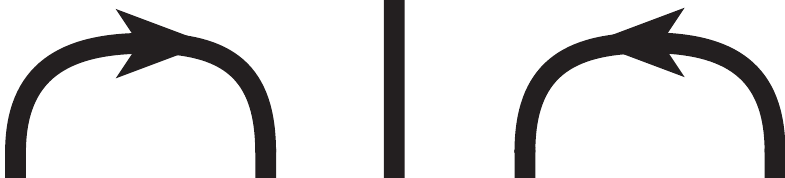,width=74pt}}\\
n\cdot {}^{-1}n\cdot s\cdot n{}^{-1}\cdot n
\end{array}
\eeq
which represent the inequalities:
\[
\raisebox{-1.5mm}{\epsfig{figure=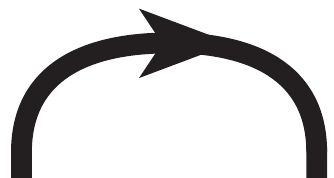,width=30pt}}\ \ \leftrightarrow  
\begin{array}{c}
1\\
\rotleq\\
n \cdot {}^{-1}n
\end{array}
\qquad\qquad\ \ \ \
\raisebox{-1.5mm}{\epsfig{figure=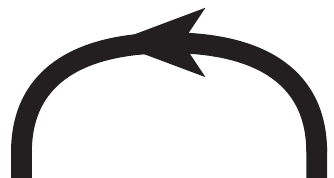,width=30pt}}\ \ \leftrightarrow  
\begin{array}{c}
1\\
\rotleq\\
n^{-1} \cdot n
\end{array}
\]
Moreover, a pregroup can be defined  in terms of cups and caps.  In category theoretic language, both the diagrammatic language for quantum axiomatics and pregroups are so-called \em compact closed categories \em \cite{KellyLaplaza, SelingerSurvey}; while  the quantum  language is \em symmetric\em,  pregroups have to be \em non-symmetric \em given the importance of word-order in sentences.

\paragraph{Solution to the NLP problem: the algorithm.} Assume a grammatically well-typed sentence and a meaning vector $\overrightarrow{v}_{\!j}$ for each of its words, which we assume to be represented in a vector space of which the tensor structure  matches the structure of its grammatical type,\footnote{How this can be achieved within the context of the vector space model of meaning is outlined   in \cite{GSCCP} and used in \cite{GS}.} e.g.:
\[
n\leadsto{\cal V}\qquad\ \  tv={}^{-1}n\cdot s\cdot n{}^{-1}\leadsto {\cal V}\otimes {\cal W}\otimes {\cal V} 
\]
where ${\cal W}$ is the vector space in which we intend to represent the meanings of sentences.  Then one proceeds as follows: 
\ben
\item Compute the tensor product $\overrightarrow{Words}=\overrightarrow{v}_{\!1}\otimes\ldots\otimes \overrightarrow{v}_{\!k}$ of the word meaning vectors in  order of appearance in the sentence; e.g.~$\overrightarrow{noun_1}\otimes \overrightarrow{verb}\otimes\overrightarrow{noun_2}$.
\item Construct a linear map $f$  that represents the type reduction as follows: given the diagram that represents a type reduction (cf.~(\ref{diagtypered}) above), we interpret caps as $\sum_i\langle ii|$ and straight wire as identities; e.g.~$\sum_i\langle ii|\otimes Id\otimes \sum_i\langle ii|$.
\item Compute $\overrightarrow{Sentence}:=f(\overrightarrow{Words})\in{\cal W}$. 
\een

Hence the crux  is: \em the grammatical correctness verification procedure becomes an actual linear map that transforms the meanings of words into the meaning of the sentence by making these words interact via caps\em.  
Does it work? The proof is in the pudding.  Proof-of-concept examples are in \cite{CSC}, and concrete experimentally verified applications are in \cite{GS}. 

We invite the reader to also look at \cite{CSC} for the example sentence: 
\begin{center}
``Alice does not like Bob'' \ ,
\end{center}
where ``does'' and ``not''
are assigned not empirical but `logical' meanings:
\[
\raisebox{-3mm}{\epsfig{figure=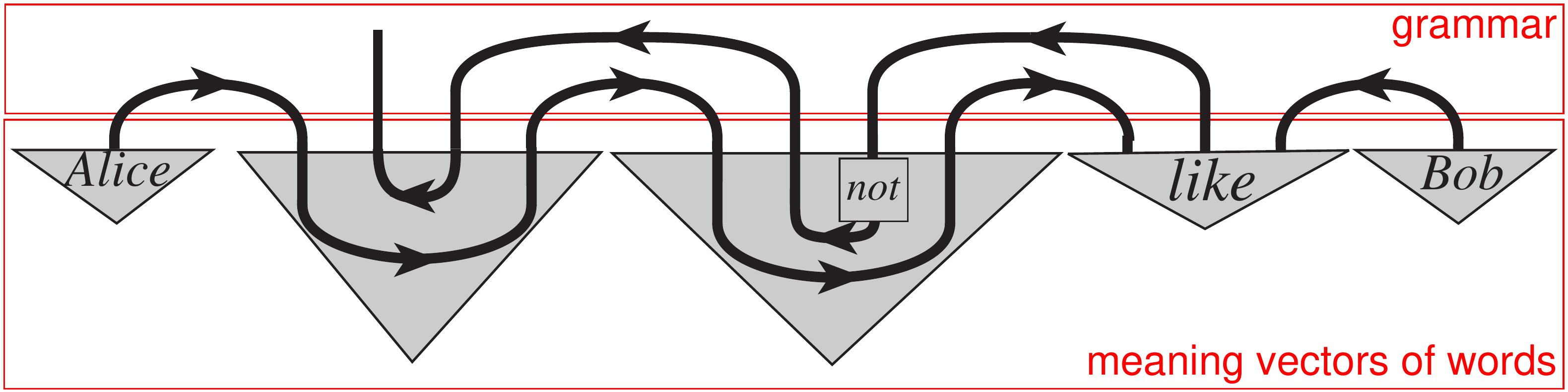,width=380pt}}
\]
resulting in a more interesting information flow structure.  In ongoing work we investigate how the structures that are used to represent classical data flow in the quantum teleportation protocol  enable one to model more of these `logical words'.  For example, in \cite{CCS2} this was done for relative pronouns such as ``who'', ``which'', ``that'' and ``whose''.

Finally, while the well-established structural similarities across
language families in terms of grammatical type calculi may seem
mysterious, the teleportation-like information flow interpretation
presented here clearly explains them.


\begin{thebibliography}{99} 

\bibitem{AC}
S.~Abramsky and B.~Coecke (2004) \em  A categorical semantics of quantum protocols\em. In: Proceedings of 19th IEEE conference on Logic in Computer Science, pages 415--425. IEEE Press. arXiv:quant-ph/0402130.  

\bibitem{BBC} 
C.~H.~Bennett, G.~Brassard, C.~Cr\'epeau, R.~Jozsa, A.~Peres and W.~K.~Wooters (1993)  \em Teleporting an unknown quantum state via dual classical and Einstein-Podolsky-Rosen channels\em.  Physical Review Letters {\bf 70}, 1895--1899.


\bibitem{BennettSchumacher}
C.~H.~Bennett and B.~Schumacher (2002) Lecture at Tata Institute for Fundamental Research, Mumbai, India.

\bibitem{Alice}
L.~Carroll (1865) \em Alice's Adventures in Wonderland\em. MacMillan and Co.

\bibitem{Chomsky}
N.~Chomsky (1957) \em Syntactic Structures\em. Mouton.

\bibitem{LE}
B.~Coecke (2004)    \em The logic of entanglement\em.   arXiv: quant-ph/0402014

\bibitem{ContPhys}
B.~Coecke (2010) \em Quantum picturalism\em. Contemporary Physics {\bf 51}, 59--83.  arXiv:0908.1787

\bibitem{NJP}
B.~Coecke and R.~Duncan (2011): \em Interacting quantum observables: categorical algebra and diagrammatics\em. New Journal of Physics {\bf 13}, 043016. arXiv:0906.4725 

\bibitem{CCS}
S.~Clark, B.~Coecke and M.~Sadrzadeh (2008) \em A compositional distributional model of meaning\em. Proceedings of AAAI Spring Symposium on Quantum Interaction, pages 133--140, AAAI Press.

\bibitem{CCS2}
S.~Clark, B.~Coecke and M.~Sadrzadeh (2013) \em The Frobenius anatomy of relative pronouns\em. Draft paper.

\bibitem{CPav}
B.~Coecke and D.~Pavlovic (2007) \em Quantum measurements without sums\em. In: Mathematics of Quantum Computing and Technology, G.~Chen, L.~Kauffman and S.~Lamonaco (eds), pages 567--604. Taylor and Francis.  arXiv:quant-ph/0608035

\bibitem{CSC}
B.~Coecke, M.~Sadrzadeh and S.~Clark (2011) \em Mathematical foundations for a compositional distributional model of meaning\em. Linguistic Analysis -- Lambek Festschrift. arXiv:1003.4394 [cs.CL]

\bibitem{Gazdar}
G.~Gazdar (1996) \em Paradigm merger in natural language processing\em. In: Computing Tomorrow: Future Research Directions in Computer Science, pages 88--109, R.~Milner and I.~Wand (eds.), Cambridge University Press.
 
 \bibitem{GSCCP}
E.~Grefenstette, M.~Sadrzadeh, S.~Clark, B.~Coecke and
S.~Pulman (2011) \em Concrete compositional sentence
spaces for a compositional distributional model of
meaning\em. In: Proceedings of the 2011 International Conference on Computational
Semantics (IWCS). arXiv:1101.0309 [cs.CL]
 
\bibitem{GS}
E.~Grefenstette and M.~Sadrzadeh (2011) \em Experimental support for a categorical compositional distributional model of meaning\em.  In: Proceedings of the 2011 Conference on Empirical Methods in Natural Language Processing (EMNLP). arXiv:1106.4058 [cs.CL]

\bibitem{Gottesman}
D.~Gottesman and I.~L.~Chuang (1999) \em Quantum teleportation is a universal computational primitive\em.  Nature {\bf 402}, 390--393.   arXiv:quant-ph/9908010

\bibitem{KellyLaplaza}
G.~M.~Kelly and M.~L.~Laplaza (1980) \em Coherence for compact closed categories\em.  Journal of Pure and Applied Algebra {\bf 19}, 193--213.

\bibitem{Lambek1}
 J.~Lambek (1958) \em The mathematics of sentence structure\em. American Mathematical Monthly {\bf 65}, 154--169.

\bibitem{Lambek2}
J.~Lambek (1999) \em Type grammar revisited\em. In: Logical Aspects of Computational Linguistics,  Lecture Notes in Artificial Intelligence   {\bf 1582}, Springer.

\bibitem{Lambek3}
 J.~Lambek (2008) \em From Word to Sentence\em. Polimetrica.


\bibitem{Montague}
R.~Montague (1974) \em Formal Philosophy: Selected Papers\em. Yale University Press. 

\bibitem{NewScientist}
New Scientist (December 11, 2010) \em Quantum links let computers understand language\em.

\bibitem{Popescu}
P.~Skrzypczyk, N.~Brunner and S.~Popescu (2009) \em Emergence of quantum correlations from nonlocality swapping\em. Physical Review Letters {\bf 102}, 110402. arXiv:0811.2937

\bibitem{VectModel}
H.~Schuetze (1998) \em Automatic word sense discrimination\em. Computational Linguistics {\bf 24}, 97--123.

\bibitem{SelingerSurvey}
P.~Selinger (2011) \em A survey of graphical languages for monoidal categories\em. In: New Structures for Physics, B.~Coecke (ed), pages 289--356. Lecture Notes in Physics 813, Springer-Verlag. 
arXiv:0908.3347

\end{thebibliography}
\end{document}